\pdfoutput=1

\documentclass[11pt]{article}

\usepackage{acl}

\usepackage{times}
\usepackage{latexsym}
\usepackage{xcolor}
\usepackage[T1]{fontenc}

\usepackage[utf8]{inputenc}

\usepackage{microtype}
\usepackage{multirow}
\usepackage{latexsym}
\usepackage{booktabs}
\usepackage{courier}
\usepackage{amsmath}
\usepackage{subfig}
\usepackage{placeins}
\usepackage{tikz}
\usepackage{hyperref}
\title{Do Transformers Encode a Foundational Ontology? \\ Probing Abstract Classes in Natural Language} 

\author{Mael Jullien$~^{\dagger}$, Marco Valentino$~^{\dagger}$$^{\ddagger}$, Andr\'e Freitas$~^{\dagger}$$^{\ddagger}$\\ Department of Computer Science, University of Manchester, United Kingdom$~^{\dagger} $\\  Idiap Research Institute, Switzerland$^{\ddagger}$\\ {\tt \{mael.jullien,marco.valentino,andre.freitas\}} \\ {\tt @manchester.ac.uk} \\}

\begin{document}
\maketitle
\begin{abstract}
With the methodological support of probing (or diagnostic classification), recent studies have demonstrated that Transformers encode syntactic and semantic information to some extent. Following this line of research, this paper aims at taking semantic probing to an abstraction extreme with the goal of answering the following research question: \emph{can contemporary Transformer-based models reflect an underlying Foundational Ontology?} To this end, we present a systematic Foundational Ontology (FO) probing methodology to investigate whether Transformers-based models encode abstract semantic information. Following different pre-training and fine-tuning regimes, we present an extensive evaluation of a diverse set of large-scale language models over three distinct and complementary FO tagging experiments. Specifically, we present and discuss the following conclusions: (1) 
The probing results indicate that Transformer-based models incidentally encode information related to Foundational Ontologies during the pre-training process; (2) Robust FO taggers (accuracy $\approx 90\%$ ) can be efficiently built leveraging on this knowledge.

\end{abstract}

\section{Introduction}

Large-scale neural language models have become the de-facto representational substrate for supporting language-based inference \cite{devlin-etal-2019-bert,robert}. More recently, these models have been used and specialised to cope with abstract inference \cite{valentino2021unification,thayaparan2021explainable} and to support efficient generalisation \cite{zhou2021encoding} in several downstream reasoning tasks. 

As large-scale language models are gradually evolving towards more abstract inference, it is crucial to study and understand the underlying semantics encoded in their representation to identify biases and inconsistencies within the models \cite{elazar2021measuring}, improve transparency \cite{thayaparan2020survey}, and further investigate their generalisation and reasoning capabilities \cite{hu2020systematic}. 

With the methodological support of probing (or diagnostic classification) \cite{elazar2021amnesic,ferreira2021does}, recent studies have demonstrated that transformers encode syntactic and semantic dependencies to some extent \cite{tenney2019bert}. The probing paradigm, in fact, has been recently employed to investigate whether abstract semantic information is encoded in large-scale language models, using auxiliary tasks such as Named Entity Recognition (NER) \cite{jin2019probing}, Semantic Role Labeling (SRL) \cite{tenney2019learn} and Semantic Annotation (SA) \cite{xu2020understanding}. 
This paper aims at taking semantic probing to an abstraction extreme with the goal of answering the following research question: \emph{can contemporary transformer-based models reflect an underlying Foundational Ontology?}

Foundational ontologies provide logical axiomatisations for domain-agnostic top level categories, such as event, biological-object or geographical-object. Foundational Ontologies contain basic and universal concepts, which are either are meta, generic, or philosophical. to promote the integration of highly general information and expressiveness over a wide range of domains \cite{schmidt2020foundational}. Their general purpose is to map a concept to its most fundamental interpretation \cite{TOPTAGGER}, e.g. \emph{``Inserted into the heart's left ventricle''}, the word \emph{``ventricle''} is mapped to \emph{Biological-Object}. Such a mapping is fundamental for enabling generalisation and reasoning since higher level categories can be adopted to deliver the required abstractive mechanisms without loss in meaning. Foundational ontologies and their ties to logics represent an important connection between natural language and reasoning \cite{TOPTAGGER}. Additionally, Foundational ontology tagging presents a task requiring a unique level of abstraction, common-sense reasoning and a deep understanding of context and polysemy.

In this work, we link the formally grounded abstract categories of Foundational Ontologies \cite{gangemi2002sweetening,guizzardi2005ontological} to contextualised embeddings by exploring two distinct research hypotheses. The first hypothesis is semantic in nature: pre-trained transformer-based models incidentally encode FO categories, and this information can be inspected through the probing paradigm. The second research hypothesis is practical: based on these pre-encoded categories, a robust lexical semantic model (a term-level FO shallow parser) can be built.

We investigate these hypotheses with the support of a systematic FO probing methodology over a diverse set of reference models, following different pre-training and fine-tuning regimes. Specifically, we introduce three different FO tagging experiments using a mixture of probing and fine-tuning techniques defined on WordNet-DOLCE alignments \cite{TOPTAGGER}, namely \emph{Basic Foundational Ontology Tagging}, \emph{Binary Task}, and \emph{Singular Contextualised Embedding Probe}. Through the definition of distinct and complementary auxiliary tasks, we are able to conduct extensive experiments to investigate the encoding of Foundational Ontologies in a diverse set of large-scale language models, including BERT \cite{devlin-etal-2019-bert} and RoBERTa \cite{robert} pre-trained on several downstream NLI datasets. The empirical evaluation resulted in the following conclusions:

\begin{itemize}
    \item The probing results indicate that transformer-based models incidentally encode information related to Foundational Ontologies during the pre-training process, with Multi-Layer-Perceptron (MLP) probes being substantially more effective in all FO probing tasks.
    \item Robust FO taggers (accuracy $\approx 90\%$ ) can be efficiently built leveraging on this knowledge.
    \item There was no significant differences between the results of different architectures or different fine-tuning procedures.
\end{itemize}

The FO probing pipeline, dataset, and tagger are all available as open source projects at the following URL: \url{anonymous-url.com}.

\section{Related Work}

\paragraph{Semantic Annotation}

Semantic annotation is a method of assigning an enumerable set of distinct classes to words, allowing for a more efficient encoding of generalisations, upon which algorithms could be applied \cite{TOPTAGGER}. The most well-documented semantic annotation tasks include: Semantic Role Labelling (SRL) \cite{larionov-etal-2019-semantic}, Sentiment annotation (SA) \cite{manning-etal-2014-stanford}, Named Entity Recognition (NER) \cite{ner} and Word Sense Disambiguation \cite{survey}. Transformer-based models and their word embeddings have been implemented to achieve state-of-the-art results on each of these tasks \cite{DBLP:journals/corr/abs-1905-05677,DBLP:journals/corr/abs-2010-05006,zhang2021semantic,senti}. 

\paragraph{Ontology Tagging}

Ontologies provide a machine-readable representation (class) of a concept in the real world \cite{bertalignment}, drawing a connection between natural language and logical reasoning \cite{TOPTAGGER}. Foundational Ontologies are a set of classes which map a concept to its most fundamental interpretation \cite{TOPTAGGER}, see examples in Table \ref{tab:examples}. Currently, there is no existing Foundational Ontology (FO) tagging dataset large enough to support extensive analysis. The most closely related task to ontology tagging is ontology alignment, the task of generating correspondences between entities contained in a set of ontologies \cite{Ardjani2015OntologyAlignmentTS}. Results from an automatic ontology alignment study indicated that BERT-based embeddings in isolation are not able to generate functional alignments \cite{neutel2021towards}. However, this study is not concerned with ontologies at a foundational level, and does not offer probing analysis.

\paragraph{WordNet-DOLCE Mapping}

The Descriptive Ontology for Linguistic and Cognitive Engineering (DOLCE) \cite{Masolo2002WonderWebDD} was created to encode categories which are inherently present within natural language and human reasoning \cite{TOPTAGGER}. WordNet \cite{fellbaum_wordnet_2012} is an English lexical database containing approximately 155,000 words, organised into an acyclic graph with semantic relations as edges and words as vertices. An alignment between WordNet upper level nouns synsets and DOLCE was proposed \cite{10.5555/958671.958673}, which was eventually expanded into fully classified verb and noun database \cite{TOPTAGGER}. This alignment will form the basis of the FO tagging task presented in this paper. \citet{TOPTAGGER} also presents Top Level Tagger (TLT), an FO annotation tool. TLT applies a rigid mapping from WordNet to DOLCE classes, this causes issues for annotating large symbolic word spaces, as any words not explicitly mapped to a DOLCE class are assigned the null label.

\paragraph{Probing NLI Models}

Probing, also denoted as auxilliary prediction \cite{adi2017finegrained} or diagnostic
classification \cite{hupkes2018visualisation}, is designed to investigate the internal representations of a model, generally to explain what information is encoded and how it influences the output \cite{belinkov2021probing}. This is done by training a basic probe model to predict a property of interest from the latent representations of the model \cite{elazar2021amnesic}.
The probing paradigm has already been applied to NER \cite{jin2019probing}, SRL \cite{tenney2019learn} and SA \cite{xu2020understanding} tasks. This paper will expand this discussion to FO categories.
However, probing is not without its limitations: there is a lack of comparative baselines and there are questions about the choice of evaluation metrics, probe models and datasets, as well as the correlational nature of the method \cite{belinkov2021probing,ravichander2021probing}. To mitigate the impact of these issues this paper will employ the Probe-Ably framework \cite{ferreira2021does} for all probing experiments, as it supports the current best practices for probing.

\begin{figure*}[t]
\centering
\includegraphics[width=0.9\textwidth]{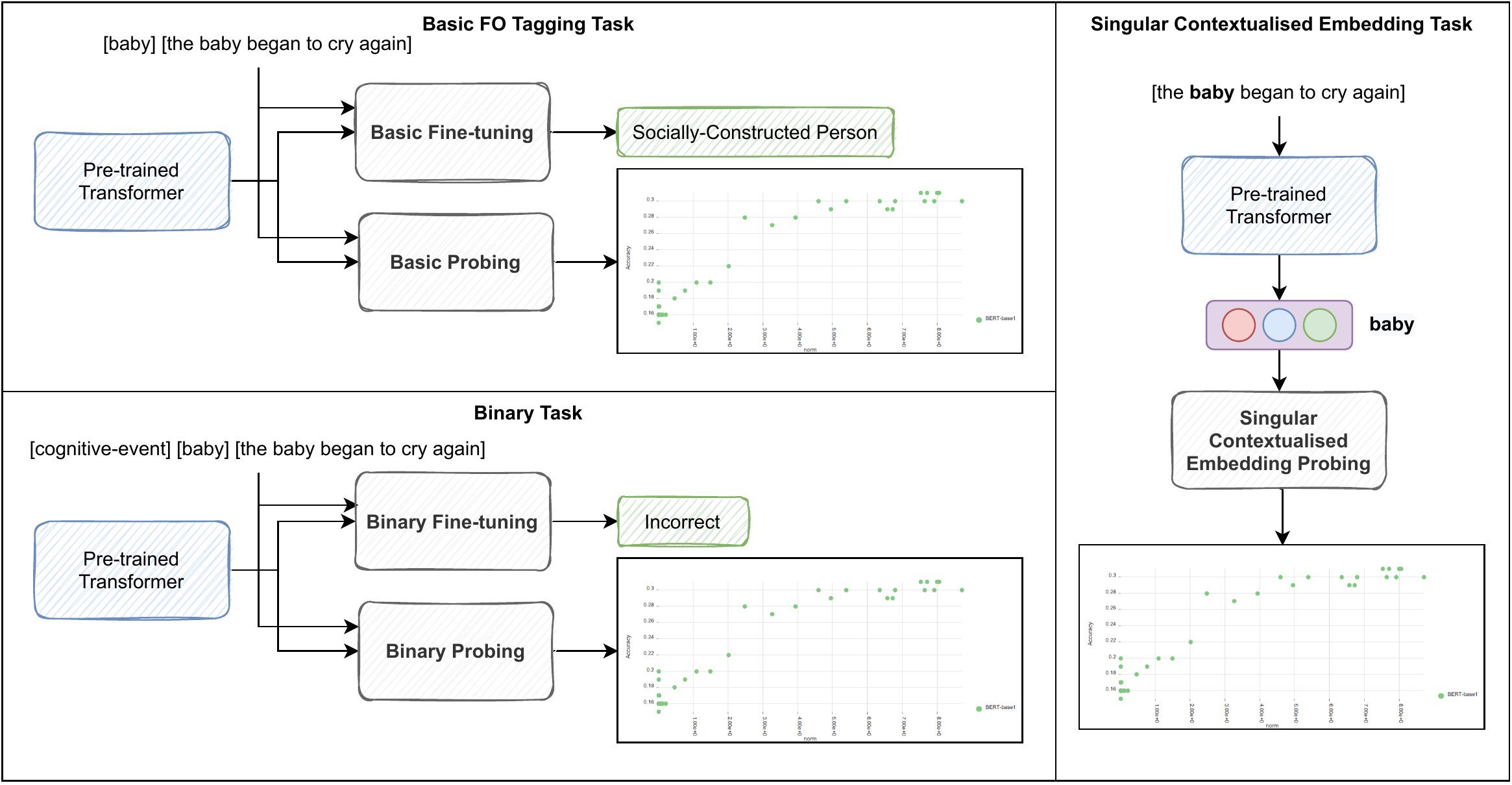}

\caption{Schematic representation of the three tasks adopted for the FO tagging probing and fine-tuning experiments.}
\end{figure*}

\section{Foundational Ontology Tagging Dataset}

The FO tagging dataset used in this paper was built by using the WordNet-DOLCE alignment \cite{TOPTAGGER}. To construct this dataset, the FO annotation tool Top Level Tagger \cite{TOPTAGGER} was used to annotate the Brown corpus \cite{francis_1965}. The Brown corpus was used, as WordNet does not consistently provide example sentences for words. The Brown corpus is comprised of a variety of English texts concerning a range of topics, totalling approximately 1 million words. The data was tagged using a subset (6) of the Top Level Tagger Foundational classes: \emph{Socially-Constructed-Person, Cognitive-Event, Geographical-Role, Biological-Object, Non-Agentive-Functional-Object} and \emph{Information-Object}, to allow for more practical manual verification. For each sentence in the corpus one tagged word was extracted at random, resulting in a data set of 2760 unique samples (460 per class), with samples in the format seen in Table \ref{tab:examples}. The Biological-Object class was supplemented with 100 samples from the WordNet alignment to ensure a sufficiently large and equally distributed set of classes. All samples in the data set were manually verified.

\section{Methodology}

The evaluation of Transformer-based model performance on the FO tagging task is guided by the following research hypotheses:
\begin{itemize}
    \item \textbf{RH1}: Pretrained Transformer-based models incidentally encode FO categories, and these can be extracted through the probing paradigm. 
    \item \textbf{RH2}: Based on these pre-encoded categories, a robust lexical semantic model (a term-level FO shallowparser) can be built.
\end{itemize}

These hypotheses will be tested on a novel Foundational Ontology tagging dataset, using an array of Transformer-based models, consisting of pretrained BERT-base \cite{devlin-etal-2019-bert}, BERT-large , RoBERTa-base \cite{robert} and RoBERTa-large architectures. Additionally, we also test on BERT-base and RoBERTa-large models fine-tuned on four benchmark NLI datasets: Multi-Genre Natural Language Inference (MultiNLI) \cite{mnli}, Stanford Natural Language Inference (SNLI) \cite{snli5}, Microsoft Research Paraphrase Corpus (MRPC) \cite{dolan-brockett-2005-automatically} and Quora Question Pairs\footnote[1]{\url{https://www.kaggle.com/c/quora-question-pairs/overview}} (QQP) (see fine-tuning results in Table \ref{tab:models}). This allows us to determine the impact of different fine-tuning tasks and architectures, as well as identifying possible correlations between model performance on the fine-tuned task and the auxiliary probe task. We define three different FO tagging experiments to test these research hypotheses: 
\begin{itemize}
    \item \textbf{Basic Foundational Ontology tagging task}: Probing and fine-tuning on a basic FO tagging task, consisting of classifying the concatenation of a given word with its respective example sentence on one of 6 FO classes.
    \item \textbf{Binary task}: Probing and fine-tuning on a Binary FO task, the classification of correct/incorrect on a concatenation of a given word with its respective example sentence and a given FO class. 
    \item \textbf{Singular contextualised embedding Probe} Probing on a FO tagging task, classifying singular contextualised word embeddings on 6 FO classes.
\end{itemize}
These 3 tasks are designed to emulate a variety of different NLP tasks, the Basic task using the same format as POS tagging, the Binary task entailment classification, and the singular contextualised embedding probe aims to force the models to encode textual context. Using this we can evaluate if the format affects performance, and if so, which format is most effective.

\label{sec:answ_sel}

\begin{table}[t]
    \centering
    \small
    \begin{tabular}{@{}p{2.5cm}p{1.3cm}p{3cm}cc@{}}
    \toprule
         \multirow{2}{*}{\textbf{Ontology Class}} &
         \multirow{2}{*}{\textbf{Word}} &
         \multirow{2}{*}{\textbf{Example sentence}}\\\\
         \midrule
         Cognitive-event &  \textbf{Trusted} & Only people he liked and \textbf{trusted} got to see.\\
         \midrule
         Socially-constructed-person &  \textbf{Baby} & The \textbf{baby} began to cry again.\\
         \midrule
         Non-agentive-functional-object &  \textbf{Car} & He needs a \textbf{car} to get to work.\\
         \midrule
        Biological-Object &  \textbf{Ventricle} & Inserted into the heart's left \textbf{ventricle}\\
         \midrule
         Information-Object &  \textbf{Appendix} & However, as noted in the \textbf{Appendix}\\
         \midrule
         Geographical-Object &  \textbf{Hemisphere} & It was night on this \textbf{hemisphere} \\
         \bottomrule
    \end{tabular}
    \caption{Example of samples from the constructed FO tagging dataset.}
    \label{tab:examples}
\end{table}

\begin{table}[t]
    \small
    \centering
    \begin{tabular}{@{}p{3.5cm}ccc@{}}
    \toprule
         \multirow{2}{*}{\textbf{NLI Pre-training task}} &
         \multicolumn{2}{c}{\textbf{Accuracy}}\\
         \cmidrule{2-3}
         &BERT-base&RoBERTa-large\\
         \midrule
         MNLI & 0.80 & 0.92 \\
         \midrule
         SNLI & 0.84 & 0.88 \\
         \midrule
         MRPC & 0.88 & 0.88 \\
         \midrule
         QQP  & 0.91 & 0.93 \\
         \bottomrule
    \end{tabular}
    \caption{Performance of the adopted language models on their respective pre-training NLI tasks}
    \label{tab:models}
\end{table}

\subsection{Basic FO Tagging Probe}

To extract the sentence embeddings, the [word][example] pairs are tokenized and input to a given model. Then the second to last hidden layer of each token is extracted from the output, the average of these layers is used as the sentence embeddings. This produces a single vector for each input data point, of size 768 for base models and 1024 for large models. 

\subsection{Basic FO Tagging Fine-tuning}

The models are fine-tuned on the FO tagging dataset using a 60-20-20 train-validation-test split, for 10 epochs, after which the checkpoint with the minimum loss is then evaluated on the test set. 

\subsection{Binary Task}

For the binary task each FO tagging dataset data point is duplicated, and the duplicate point is randomly assigned an incorrect FO class. The original point is given an additional \emph{Correct} label and the duplicate is labelled \emph{Incorrect}. The result is a binary data set containing 5520 samples with a 1 to 1 ratio of correct/incorrect FO labels. The premise hypothesis pairs for this task are constructed by introducing a new special token (TSEP) to the model's tokenizer, to create the premise, [word][TSEP][example sentence], and the hypothesis, [FO class]. Passed to the model as [FO class][SEP][word][TSEP][example sentence], for binary classification. It should be noted that none of the models were pre-trained with this new token. 

\subsubsection{Binary Probe}

The sentence embeddings for the binary samples are extracted using the same process as in the basic FO probe, as an average of the second to last hidden layer for each token.

\subsubsection{Binary Fine-tuning}

The models are fine-tuned on the binary dataset using a 60-20-20 train-validation-test split, for 10 epochs, after which the checkpoint with the minimum loss is then evaluated on the test set.

\subsection{Singular Contextualised Embedding}

The auxiliary task for this experiment is the classification of singular contextualised word embeddings on 6 FO classes. To generate a singular contextualised word embedding from a [word][example] pair, the example sentence is tokenized and input to a given model, and for each token generated from the target word (i.e. for the target word \emph{backboard}, the resulting tokens are [back, \#\#board]) the last 4 hidden layers are concatenated together. These are then averaged out to form the singular contextualised word embedding. For base models these embeddings have a size of 3072, and for large models a size of 4096. The same precautions are applied here as in the basic FO tagging probe.

\begin{table*}[t]
\resizebox{\textwidth}{!}{
\begin{tabular}{@{}llllllllllllllll@{}}
\toprule
& \multicolumn{4}{c}{\textbf{RoBERTa Large}} & \multicolumn{1}{c}{}& \multicolumn{1}{c}{} & \multicolumn{1}{c}{}& \multicolumn{4}{c}{\textbf{BERT base}} & \multicolumn{1}{c}{} & \multicolumn{1}{c}{} & \multicolumn{1}{c}{} \\
\cline{2-6} \cline{8-12}  &MNLI & SNLI & MRPC& QQP   & base &  &MNLI & SNLI  & MRPC& QQP & base & & BERT Large  & RoBERTa base \\ \midrule
\textbf{Basic FO tagging task Accuracy}         & 0.89 & 0.89 & 0.89   & \textbf{0.90}& \textbf{0.90} &   & 0.88 & 0.87   & 0.88 & 0.89 & 0.89  &   &\textbf{0.90}   & 0.86   \\\midrule  

\textbf{Binary task Accuracy}        & 0.89 & 0.89  & 0.89   & \textbf{0.90} & 0.89 &   & \textbf{0.90} & 0.89   & \textbf{0.90} & \textbf{0.90} & \textbf{0.90}& & 0.89 & 0.89 \\\bottomrule \\
\end{tabular}}

\caption{Results for Basic Foundational Ontology Tagging task and Binary task on the test set \textbf{(fine-tuning)}.}
\label{tab:base fine}
\end{table*}

\begin{table}[t]
\centering
\resizebox{\columnwidth}{!}{%
\begin{tabular}{@{}lllll@{}}
\toprule
                & \multicolumn{2}{c}{\textbf{MLP probe}} & \multicolumn{2}{c}{\textbf{Linear probe}}\\
\textbf{Model}  & Accuracy & Max Selectivity & Accuracy & Max Selectivity\\ \midrule
Random baseline       & 0.17          & -              & 0.17        & -       \\\midrule\midrule
BERT base       & 0.54          & 0.27              & \textbf{0.44}          & \textbf{0.25}           \\\midrule
BERT MNLI       & 0.55          & 0.18              & 0.36          & 0.16    \\\midrule
BERT SNLI       & 0.53            & 0.20            & 0.36          & 0.14       \\\midrule
BERT QQP        & 0.54          & 0.22              & 0.41         & 0.22      \\\midrule
BERT MRPC       & 0.42            & 0.25            & 0.42         & 0.21       \\\midrule
BERT Large      & 0.47          & 0.22            & 0.39         & 0.22   \\\midrule\midrule
RoBERTa base             & 0.53        & 0.28           & 0.37          & 0.16         \\\midrule
RoBERTa Large MNLI      & 0.34      & 0.14         & 0.18          & 0.04            \\\midrule
RoBERTa Large SNLI       & 0.50       & 0.16             & 0.28          & 0.10       \\\midrule
RoBERTa Large QQP       & 0.28       & 0.06             & 0.26         & 0.05             \\\midrule
RoBERTa Large MRPC       & 0.41       & 0.21             & 0.37         & 0.18           \\\midrule
RoBERTa Large  & \textbf{0.57} & \textbf{0.30}     & 0.40          & 0.20  \\\bottomrule \\
\end{tabular}%
}

\caption{Basic FO tagging results \textbf{(probing)}.}
\label{tab:base probe}
\end{table}

\begin{table}[t]
\centering
\resizebox{\columnwidth}{!}{%
\begin{tabular}{@{}lllll@{}}
\toprule
                & \multicolumn{2}{c}{\textbf{MLP probe}} & \multicolumn{2}{c}{\textbf{Linear probe}}\\
\textbf{Model}  & Accuracy & Max selectivity & Accuracy & Max selectivity\\ \midrule
Random baseline       & 0.5          & -              & 0.5        & -             \\\midrule\midrule
BERT base       & 0.64           & 0.12         & 0.61         &0.09\\\midrule
BERT MNLI       & 0.59          & 0.06          & 0.50      & 0.03\\\midrule
BERT SNLI       & \textbf{0.66}       & 0.12       & \textbf{0.63}         & \textbf{0.10}         \\\midrule
BERT QQP       & 0.59       & 0.06             & 0.57        & 0.04            \\\midrule
BERT MRPC       & 0.65       & 0.12             & 0.61         & 0.08           \\\midrule
BERT Large      & 0.64       & \textbf{0.14}             & 0.61         & 0.08            \\\midrule\midrule
RoBERTa base   & 0.62       & 0.05             & 0.59         & 0.04    \\\midrule
RoBERTa Large  MNLI    & 0.66         & 0.10          & 0.57  & 0.05   \\\midrule
RoBERTa Large SNLI       & 0.65       & 0.08             & 0.59         & 0.05          \\\midrule
RoBERTa Large QQP       & 0.58       & 0.07             & 0.56         & 0.05          \\\midrule
RoBERTa Large MRPC       & 0.63       & 0.10             & 0.56         & 0.01  \\\midrule           
RoBERTa Large    & 0.64&  \textbf{0.14} & 0.56      & 0.06   \\\bottomrule 
\end{tabular}%
}

\caption{Binary task results \textbf{(probing)}. }
\label{tab:bin probe}
\end{table}

\section{Empirical Evaluation}

The two fine-tuning experiments were run on the same hyper-parameters, 10 epochs, using a training batch size of 16, an evaluation batch size of 64, 100 warm up steps, a weight decay of 0.01 and seed 42. The fine-tuned models are evaluated on their accuracy on the test set.  For the probing experiments this paper will employ the Probe-Ably framework \cite{ferreira2021does}, an extendable probing framework designed to streamline the process of running probing experiments according to the best practices for probing. For each probe we record the maximum selectivity achieved, and the associated accuracy. The supplementary materials contains the Probe-Ably visualisations for all linear and MLP probes, as well as the source code for the probing pipeline.

\subsection{Probing Configuration}

For each individual probe we adopt both linear and MLP models. We test each architecture on 4 different transformer-based-models, BERT base, BERT large, RoBERTa base and RoBERTa large. Additionally, we test with BERT base and RoBERTa large models fine-tuned on MNLI, SNLI, QPP and MRPC. We train 50 probes of fluctuating complexities for 25 epochs on the Probe-Ably Framework, using the default train-validation-test split and a batch size of 128.

\subsection{Probe Complexity}

The Probe-Ably Framework initiates each probe model using an approximate complexity which is varied in a controlled manner across the 50 probe models \cite{ferreira2021does}. The variance in complexity is introduced as a means to alleviate the effect of overly expressive probes simply over fitting to the auxiliary task \cite{pamento}.

Probe-Ably employs the same default hyperparameter ranges as \cite{piment2} for the MLP probes and MLP complexity is approximated using the hidden size of intermediate layers. 

Probe-Ably follows the convention proposed in \cite{pamento} for approximating and controlling linear probe model complexity. That is for $ \hat{y} = \textbf{W} \mathbf{x} + \mathbf{b} $, the nuclear norm (Equation 1) of the transformation matrix \textbf{W}, is utilised to approximate linear model complexity. The theory behind this is that the nuclear norm is an approximation of rank(\textbf{W}).

\begin{equation}
\small
     ||\mathbf{W}||_{*} = \sum_{i=1}^{min(|\mathcal{T}|, d)}\sigma_i(\mathbf{W}).
\end{equation}

The variance of the linear model complexity is regulated through the cross-entropy loss function used in the probe's training loop, via a parameter $\lambda$ (Equation 2).

\begin{equation}
\small
    -\sum_{i=1}^{n}\log p(t^{(i)} \mid \mathbf{h}^{(i)}) + \lambda \cdot ||\mathbf{W}||_{*}  
\end{equation}

\subsection{Metrics and Control Tasks}

Probing models, particularly large MLP models can be susceptible to over-fitting to noise within data \cite{ferreira2021does}, potentially even fitting to entirely random noise \cite{zhang-bowman-2018-language}. Therefore, recording a high probe model accuracy on the auxiliary task test set is not enough to confirm the presence of encoded information. Hence, selectivity is introduced, defined as the difference between a probe models accuracy on a control task and the auxiliary task \cite{hewit}. By carefully monitoring selectivity we are able to assess how much a probe model is fitting to noise, or genuine patterns in data. Probe-Ably generates control tasks by assigning random labels to the auxiliary data. 
It should be noted that that higher dimensional embeddings often require a higher probe complexity to achieve maximum accuracy than their lower dimension counterparts \cite{ferreira2021does}. This can potentially result in a superficially worse accuracy. This discrepancy has been accounted for in each task by ensuring that the probing hyper parameters allow larger models to reach, and pass a clear apex in selectivity. For each probe, we record the maximum selectivity achieved, and the accuracy produced at that same probe complexity. Additionally, we provide a graph of the results for each probe in the supplementary materials.

\begin{figure*}[t]
\centering
\subfloat[Accuracy \label{fig:map_exp_length}]{\includegraphics[width=0.625\textwidth]{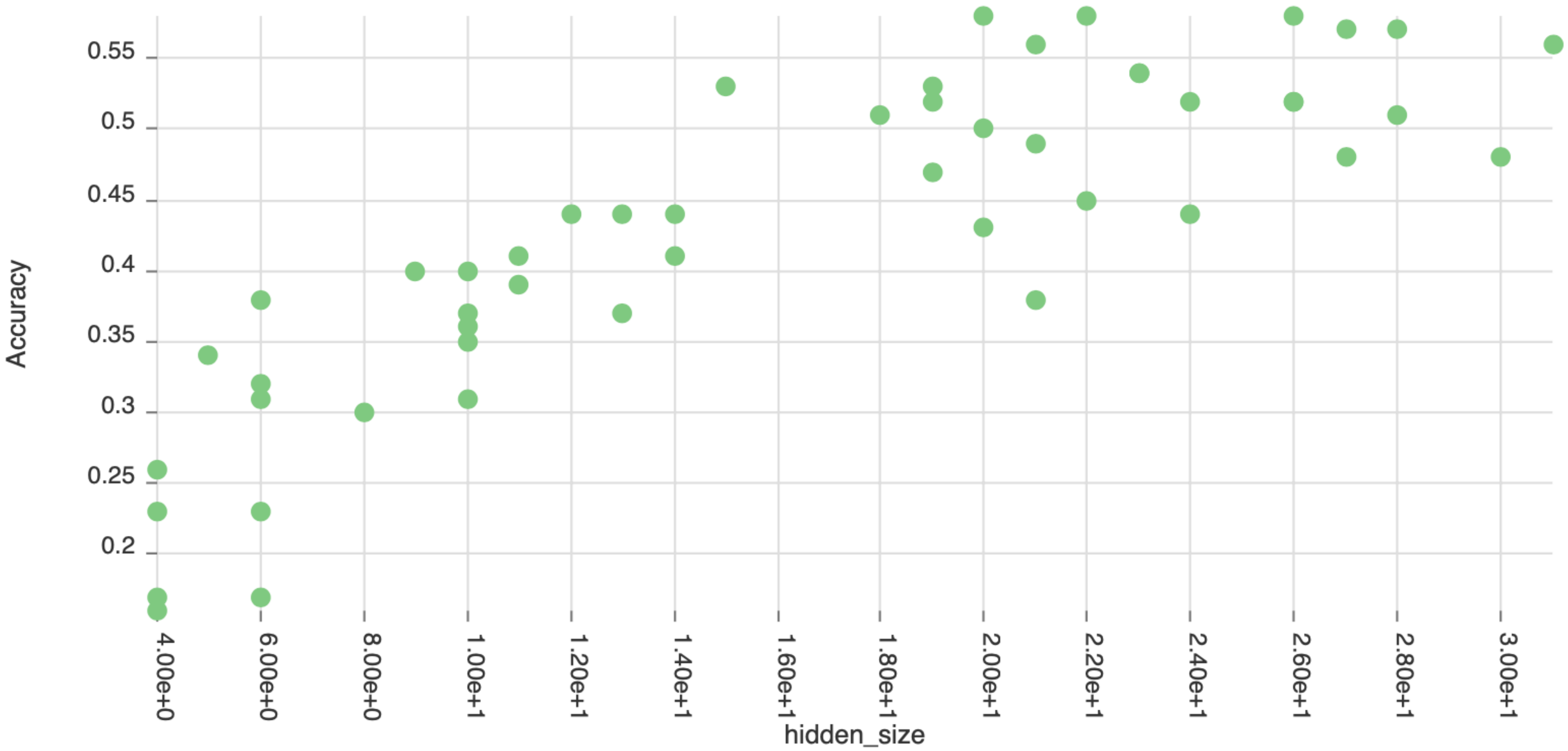}}
\hfill
\subfloat[Selectivity \label{fig:precision_k}]{\includegraphics[width=0.625\textwidth]{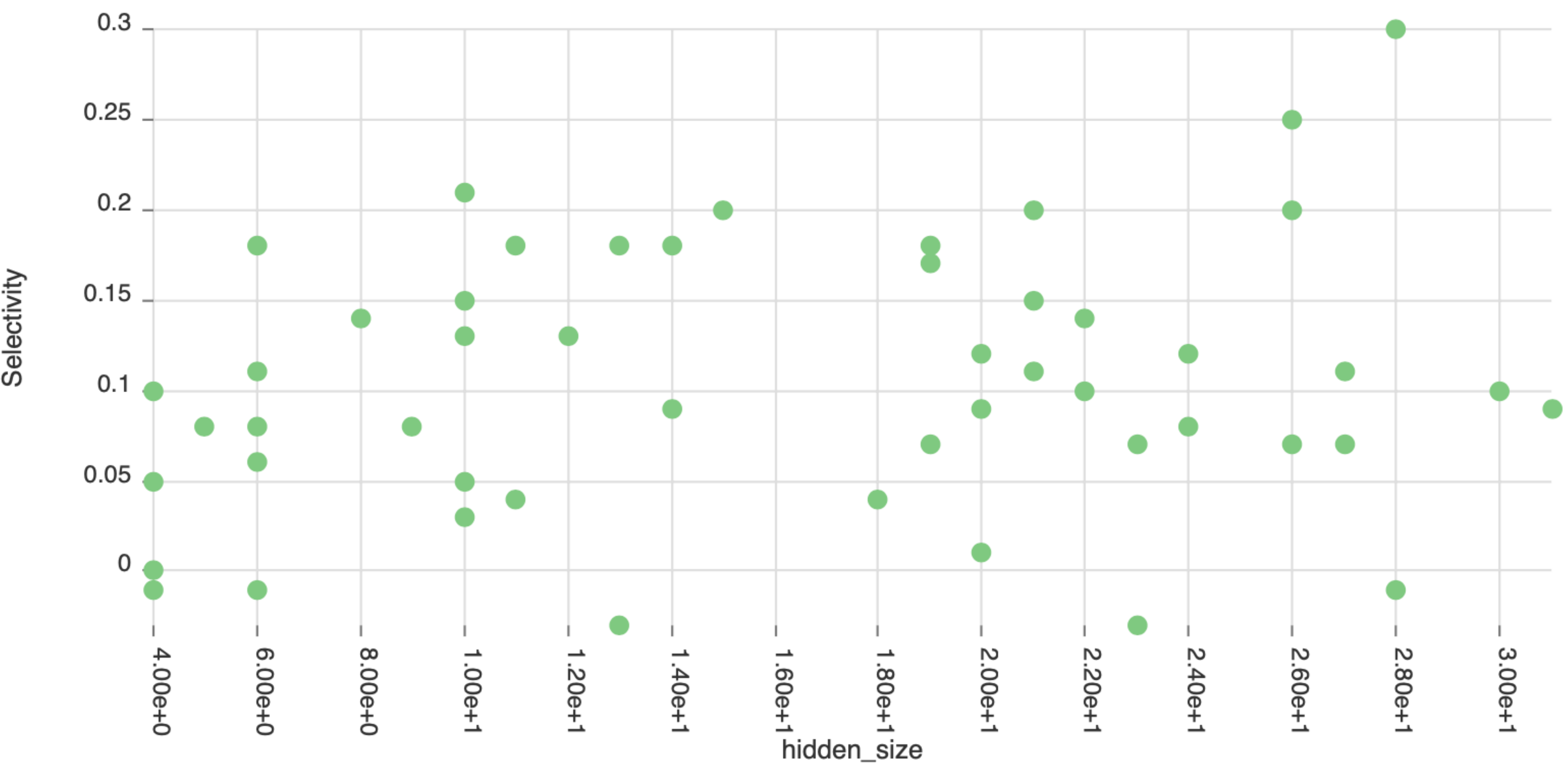}}
\caption{RoBERTa Large MLP results for the basic Foundational Ontology tagging task \textbf{(probing)}.}
\label{fig: base probe}
\end{figure*}

\begin{table}
    \centering
    \small
    \begin{tabular}{@{}ll@{}}
    \toprule
         \multirow{2}{*}{\textbf{Label}} &
         \multirow{2}{*}{\textbf{Accuracy}}\\\\
         \midrule
Socially-constructed-person  & $\textbf{0.946}$ \\
\midrule
Non-agentive-functional-object  & $0.878$ \\
\midrule
Cognitive-event  & $0.915$ \\
\midrule
Information-Object  & $0.863$ \\
\midrule
Geographical-Object  & $0.873$ \\
\midrule
Biological-Object  & $0.857$ \\
\bottomrule
    \end{tabular}
    \caption{Roberta-large class-wise accuracy on the Foundational Ontology tagging test set \textbf{(fine-tuning)}.}
    \label{tab:qualitative_examples}
    \label{tab:class base}
\end{table}

\begin{table}
    \centering
    \small
    \begin{tabular}{@{}ll@{}}
    \toprule
         \multirow{2}{*}{\textbf{Label}} &
         \multirow{2}{*}{\textbf{Accuracy}}\\\\
         \midrule
Incorrect  & $0.864$ \\
\midrule
Correct  & $\textbf{0.940}$ \\
\bottomrule
    \end{tabular}
    \caption{BERT-base class-wise accuracy on the binary tagging test set \textbf{(fine-tuning)}.}
    \label{tab:bin class}
\end{table}

\begin{table}[t]
\resizebox{\columnwidth}{!}{%
\begin{tabular}{@{}lllll@{}}
\toprule
                & \multicolumn{2}{c}{\textbf{MLP probe}} & \multicolumn{2}{c}{\textbf{Linear probe}}\\
\textbf{Model}  & Accuracy & Max selectivity & Accuracy & Max selectivity \\ \midrule

Random baseline       & 0.17          & -              & 0.17        & -              \\\midrule\midrule
BERT base       & 0.64 & 0.26      & 0.49           & 0.30               \\\midrule
BERT MNLI       & 0.64           & 0.25               & 0.54          & 0.22               \\\midrule
BERT SNLI       & 0.54       & 0.22             & 0.58         & 0.35              \\\midrule
BERT QQP       & 0.50       & 0.17             & 0.55         & 0.34              \\\midrule
BERT MRPC       & 0.55       & 0.25             & 0.56         & 0.35              \\\midrule
BERT Large      & 0.53       & 0.08             & 0.39         & 0.20              \\\midrule\midrule
RoBERTa base   & 0.55       & 0.24             & 0.50         & 0.20    \\\midrule
RoBERTa Large  MNLI    & 0.34          & 0.10               & 0.27           & 0.08   \\\midrule
RoBERTa Large SNLI       & \textbf{0.66}       & \textbf{0.27}             & 0.47         & 0.18            \\\midrule
RoBERTa Large QQP       & 0.35       & 0.16             & 0.28         & 0.10             \\\midrule
RoBERTa Large MRPC       & 0.48       & 0.25             & \textbf{0.62}         & \textbf{0.37}  \\\midrule    
RoBERTa Large   & 0.55          & 0.32               & 0.56          & 0.12   \\\bottomrule \\
\end{tabular}%
}

\caption{Singular contextualised word embedding results \textbf{(probing)}}
\label{tab:sing probe}
\end{table}

\subsection{Basic FO Tagging Fine-tuning}

All models effectively learned the task, with a minimum accuracy of 0.86 (RoBERTa base), results can be seen in Table \ref{tab:base fine}. The maximum accuracy recorded was 0.90 by Bert large, RoBERTA large and RoBERTa QQP. Table \ref{tab:class base} shows a class-wise error breakdown for RoBERTa large on the test set. This revealed an imbalanced model performance, with a maximum accuracy of 0.946 on the socially-constructed-person class, and a minimum of 0.857 on the biological object class (0.089 differential). The large architectures each outperformed their base counterparts, with a +0.01 difference between the best large and base models. Additionally, the models fine-tuned on NLI tasks  performed slightly worse on average, than the base versions, with a maximum -0.02 disparity.

\subsection{Binary Fine-tuning}

All models effectively learned the binary task, with a minimum accuracy of 0.89, and a maximum on 0.90, achieved by  results can be seen in Table \ref{tab:base fine}. Table \ref{tab:bin class} shows a class-wise error breakdown for BERT base on the test set, again a significant imbalance in performance can be observed as the model achieves a 0.940 accuracy on the Correct class and only 0.864 on the Incorrect class (-0.076 differential). There was very little difference in performance across model types and fine-tuning procedures.

\subsection{Basic FO Tagging Probe}

The Basic FO tagging dataset contains 2,760 samples, 460 per class. During probing this was split 552 - 552 - 1,656 as train-validation-test. The results from fine-tuning task on the basic FO tagging Probe can be found in Table \ref{tab:base probe}. 
All of the probes, apart from the RoBERTa large MNLI linear probe, significantly outperform the random baseline accuracy ($0.17$ with 6 classes) at max selectivity. The RoBERTa large MNLI linear probe produces the lowest accuracy 0.18, as well as the lowest max selectivity, 0.04. The RoBERTa large MLP probe generated the highest accuracy at max selectivity, +0.01 accuracy +0.03 max selectivity above next best probe, results seen in FIGURE \ref{fig: base probe}. MLP probes also significantly outperformed the linear probes on this task, with a +0.13 accuracy +0.05 max selectivity difference between their best probes, RoBERTa large and BERT base respectively. The models fine-tuned on NLI benchmarks all performed worse than their base versions on both probe types. The most significant difference was recorded between RoBERTa large and RoBERTa large MNLI, with a +0.23 accuracy and +0.16 max selectivity differential. With regards to the results from the 4 architectures, the largest discrepancy in the base models was between RoBERTa large and BERT large with +0.10 accuracy and +0.08 max selectivity.

\subsection{Binary Probe}
The Basic FO tagging data set contains 5,520 samples, 2,760, per class. During probing this was split 1,104 - 1,104 - 3,312 train-validation-test. Table \ref{tab:bin probe} shows the results of the binary probe. The graphs of the results for each probe are provided in the supplementary materials
All of the probe models, aside from BERT MNLI, outperform the random baseline accuracy ($0.5$ with two classes). The minimum accuracy at max selectivity was recorded on the BERT MNLI model, at 0.5, for the Linear probe type. The lowest accuracy at max selectivity for the MLP probe was produced by RoBERTa large QQP, at 0.58. The best performing models were BERT SNLI, RoBERTa large and BERT large, with BERT SNLI recording the highest accuracy, 0.66, and RoBERTa large and BERT Large generating the highest max selectivity, 0.14. As performance is evaluated on a combination of accuracy and maximum selectivity, it is difficult to make definitive claims about which of these probes was most effective. The MLP probes outperform each of their Linear counterparts, with the difference between their respective best results being +0.03 accuracy and +0.04 max selectivity. The most significant difference between a model and its NLI fine-tuned version occurs with the BERT base and BERT MNLI linear probes with +0.11 accuracy and +0.06 max selectivity. On the MLP probes BERT and RoBERTa large both achieve 0.64 accuracy and 0.14 max selectivity, in comparison, the weakest performing base model, RoBERTa base, records at 0.62 accuracy and 0.05 max selectivity.

\subsection{Singular Contextualised Embedding Probe}

The Singular contextualised embedding data set contains 2,760 samples, 460, per class. During probing this was split 552 - 552 - 1,656 train-validation-test. The results from singular contextualised embedding probe be found in Table \ref{tab:sing probe}. 
All probes significantly outperform the random baseline accuracy ($0.17$ with 6 classes). The highest accuracy, 0.66, was produced by the RoBERTa large SNLI MLP probe at 0.27 max selectivity. However, the highest max selectivity, 0.37, was generated by RoBERTa large MRPC linear probe, with a 0.62 accuracy. For this task several of the linear probes outperformed their MLP counterparts, BERT SNLI, QPP and MRPC, aswell as well as RoBERTa MRPC and RoBERTa large. The difference between the best MLP and Linear probes was +0.04 accuracy and -0.10 Max selectivity. There was no significant difference between the base models and their best NLI fine-tuned equivalent. With regards to different architectures, the best base architecture was BERT base with 0.64 accuracy and 0.26 max selectivity, and the worst was BERT large with 0.53 accuracy and 0.08 selectivity.

\section{Discussion}

\subsection{Fine-tuning Experiments}

The results from the fine-tuning on both the binary and basic FO tagging tasks clearly show that transformer based models have a propensity for encoding FO information, achieving a maximum of 0.90 accuracy with several models, on both tasks. Class-wise error breakdown revealed a significant variation in accuracy across classes for both the binary task and the basic FO tagging task. Models fine-tuned on NLI benchmarks performed consistently worse than the base models. Additionally, RoBERTa large models outperformed BERT on the  task, and the reverse was true on the binary task. However, as the test set for the basic FO task only contained 552 samples, and 1,104 for the binary task, the discrepancies are not significant enough to make any definitive statements with regards to either the effect of NLI fine-tuning or differing architectures.

\subsection{Probing Experiments}

For each of the probing experiments the majority of the probes achieved a significantly higher accuracy than the random baseline, indicating that the transformer-based models which were tested incidentally encode some FO information during their pre-training. The MLP probes heavily outperformed the linear probes for the base and binary tasks, and were closely matched in the singular contextualised embedding task. However, the MLP probes generated the highest accuracy for all tasks. With regards to the 4 different base architectures and the NLI fine-tuned models, there was no significant differences in the results. The complete set of probe visualisations are available in the supplementary materials.

\section{Conclusion}

In this work, we aim to answer the question: \emph{can contemporary transformer-based models reflect an underlying Foundational Ontology?} We present a novel FO tagging data set, and apply a systematic FO probing methodology on a range of transformer-based models with varied pre-training and fine-tuning protocols. An extensive evaluation of the probes demonstrated that transformer-based models incidentally encode information related to Foundational Ontologies during their pre-training. In particular, MLP probes are significantly more effective than linear probes in all probing tasks, and there was no significant differences between the results of different architectures or fine-tuning procedures. Additionally, it showed the effectiveness of training transformer based models on the FO dataset to produce Robust term-level FO taggers, opening the way for future research into sequence-to-sequence taggers for abstract ontology classes.

\bibliography{eacl2021}
\bibliographystyle{acl_natbib}

\appendix

\section{Supplementary Material}

\subsection{Data and code}
All the models used in the paper can be found on URL: \url{https://huggingface.co/}. The full data set along with the graphs for all of the probe results and the code to reproduce the experiments described in the paper is available at the following URL: \url{anonymous-url.com}

\end{document}